\documentclass[11pt]{article}

\usepackage[preprint]{acl}

\usepackage{pifont}
\newcommand{\cmark}{\ding{51}}
\newcommand{\xmark}{\ding{55}}
\usepackage{hyperref}
\usepackage{times}
\usepackage{latexsym}
\usepackage{tabularx}
\usepackage{amsmath}
\usepackage[T1]{fontenc}
\usepackage[table,xcdraw]{xcolor} 
\usepackage[utf8]{inputenc}
\usepackage{multirow}
\usepackage{booktabs}
\usepackage{microtype}
\usepackage{enumitem}

\usepackage{inconsolata}

\usepackage{graphicx}
\usepackage[most]{tcolorbox}
\usepackage{colortbl}
\definecolor{prompt_yellow}{RGB}{249,240,200}
\definecolor{prompt_green}{RGB}{219,231,231}
\definecolor{prompt_blue}{RGB}{186,216,240}
\definecolor{prompt_purple}{RGB}{219,205,240}
\definecolor{frame1}{RGB}{111,135,141}
\definecolor{frame2}{RGB}{69,82,129}
\definecolor{frame3}{RGB}{121,117,99}
\definecolor{frame4}{RGB}{115,93,120}
\definecolor{model_type}{RGB}{224,225,221}

\title{LongInsightBench: A Comprehensive Benchmark for Evaluating Omni-Modal Models on Human-Centric Long-Video Understanding}

\author{
  ZhaoYang Han$^{1}$ \quad
  Qihan Lin$^{1}$ \quad
  Hao Liang$^{2}$ \quad
  Bowen Chen$^{1}$ \quad
  Zhou Liu$^{2}$ \quad
  Wentao Zhang$^{2}$ \\
  \\
  $^{1}$Huazhong University of Science and Technology \\
  $^{2}$Peking University \\
  \texttt{\{zyhan04, qh\_lin, mchust\}@hust.edu.cn} \\
  \texttt{\{hao.liang, zhouliu25, wentao.zhang\}@pku.edu.cn} \\
}

\begin{document}
\maketitle
\begin{abstract}
We introduce \textbf{LongInsightBench}, the first benchmark designed to assess models' ability to understand long videos, with a focus on human language, viewpoints, actions, and other contextual elements, while integrating \textbf{visual, audio, and text} modalities. Our benchmark excels in three key areas: \textbf{a) Long-Duration, Information-Dense Videos:} We carefully select approximately 1,000 videos from open-source datasets FineVideo based on duration limit and the information density of both visual and audio modalities, focusing on content like lectures, interviews, and vlogs, which contain rich language elements. \textbf{b) Diverse and Challenging Task Scenarios:} We have designed six challenging task scenarios, including both Intra-Event and Inter-Event Tasks. \textbf{c) Rigorous and Comprehensive Quality Assurance Pipelines:} We have developed a three-step, semi-automated data quality assurance pipeline to ensure the difficulty and validity of the synthesized questions and answer options. Based on LongInsightBench, we designed a series of experiments. Experimental results shows that Omni-modal models(OLMs) still face challenge in tasks requiring precise temporal localization (T-Loc) and long-range causal inference (CE-Caus). Extended experiments reveal the information loss and processing bias in multi-modal fusion of OLMs. Our dataset and code is available at \url{https://anonymous.4open.science/r/LongInsightBench-910F/}.

\end{abstract}

\section{Introduction}
The rapid progress of large pre-trained models has advanced multimodal understanding to the forefront of artificial intelligence research. 
While Vision-Language Models (VLMs) \cite{bai2025qwen25vltechnicalreport, bai2023qwenvl, radford2021CLIP} and Audio-Language Models (ALMs) \cite{huang2023audiogptunderstandinggeneratingspeech, radford2022whisper} excel at short clips and speech, recent \textbf{Omni-modal Models (OLMs)} aim for unified perception across all modalities. 
However, evaluation remains limited: current benchmarks fail to assess a model’s ability to comprehend complex, continuous, and multimodal information in \textbf{long videos}.



\begin{figure}[t]
     \centering
     \includegraphics[width=1\linewidth]{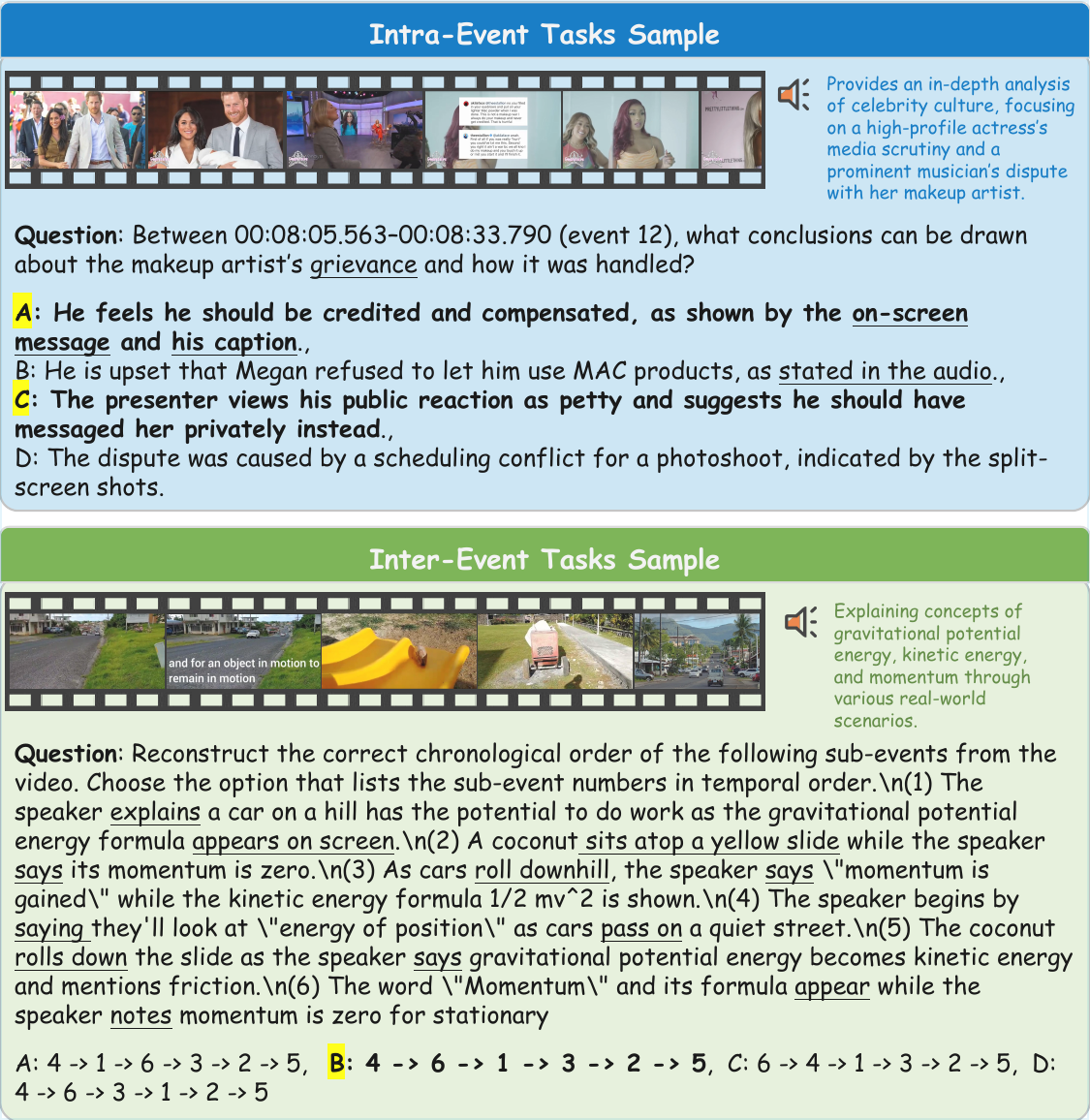}
     \caption{\textbf{Task Samples in LongInsightBench.} The upper one comes from IE-Rea(Intra-Event Reasoning) subcategory and the lower one comes from T-Recon(Timeline Reconstruction) subcategory.}
    \label{fig:category_result}
    \vspace{-0.2in}
\end{figure}

Existing datasets such as MSR-VTT \cite{xu2016msr}, ActivityNet \cite{heilbron2015activitynet}, and Ego4D \cite{grauman2021ego4d} focus on short-term tasks like action recognition or clip-based QA, requiring only local perception and reasoning within limited time windows. 
In contrast, real-world long-form content (such as lectures, interviews and vlogs) often spans tens of minutes and carries dense cross-modal information. Understanding such content demands \textbf{long-range temporal dependency modeling}, \textbf{precise cross-modal alignment and fusion} (especially between spoken language and visual context), and deep comprehension of \textbf{subtle contextual elements} \cite{videorag2025}.



Long videos are central to human communication. 
Effective comprehension requires perceiving cues that are deeply interwoven with language: tracking a speaker’s evolving \textbf{viewpoint}, discerning shifts in \textbf{sentiment}, interpreting nuanced \textbf{actions}, and understanding the \textbf{intent} behind spoken words \cite{you2024fdvs}. 
Yet, current benchmarks remain limited in three aspects:  
(1) \textbf{Duration}, lack of evaluation for maintaining attention and contextual coherence beyond a few minutes;
(2) \textbf{Modality}, neglect the critical role of the audio modality and underuse of rich linguistic information presented in long videos; 
(3) \textbf{Reasoning depth}, test surface matching only rather than distinguish deep, cross-event reasoning \cite{feng2025vbencheng}.


To address these gaps, we introduce \textbf{LongInsightBench}, the first benchmark explicitly designed for long-video omni-modal understanding. 
It emphasizes human-centered cues such as viewpoint, intent and actions, while integrating visual, audio and textual modalities for holistic multimodal understanding and reasoning.

LongInsightBench comprises a curated set of around 1,000 high-density long videos selected from FineVideo \cite{Farré2024FineVideo} and 6 challenging task types that span the spectrum of reasoning complexity. A semi-automated quality control pipeline ensures that each question requires genuine multi-modal reasoning rather than simple retrieval or single-modality clues.


We further perform systematic evaluations of state-of-the-art OLMs, complemented by additional experiments with VLMs, ALMs, and LLMs, as well as ablation studies examining the impact of video frame sampling.
The results reveal a clear performance hierarchy: large proprietary models such as Gemini2.5 excel in global comprehension but still struggle with temporal localization and long-range causal inference. 
Our analysis also identifies a persistent \textit{fusion deficit}, where multimodal integration leads to information loss and bias. 
Ablation findings show that denser frame sampling generally enhances accuracy, though the improvement varies across models.

In summary, we have three main contributions: 
\begin{enumerate}[itemsep=-1pt, topsep=2pt, leftmargin=1.5em]
    \item We introduce \textbf{LongInsightBench}, the first comprehensive benchmark for long-video omni-modal understanding, featuring about 1,000 linguistically rich long videos.
    \item We have conducted extensive \textbf{evaluations} on LongInsightBench, which establish a clear performance hierarchy among OLMs and highlights challenges in temporal localization and long-range causal inference.
    \item We reveal a consistent \textbf{\textit{fusion deficit}} in current OLMs, providing insights into limitations of existing multi-modal fusion mechanisms.
\end{enumerate}



\section{Related Works}
\subsection{Multimodal Large Language Models}
Recent multimodal large language models (MLLMs) integrate text, vision, and audio for unified reasoning.
Early efforts focused on vision–language alignment \cite{radford2021CLIP, jia2021ALIGN}, later extending to video understanding \cite{zhang2023videollama, Maaz2023VideoChatGPT} and audio perception \cite{girdhar2023imagebind, bai2023qwenvl}.
Modern omni-modal models employ modular encoders for each modality \cite{Wang2025InternVL3.5AO,vteam2025glm45v}, achieving unified inference but still struggling with temporal alignment and deep cross-modal fusion.
Recent advances improve zero-/few-shot performance \cite{GeminiTeam2025Gemini2P} and enable fine-grained visual reasoning \cite{wu2025qwenimage}, motivating further research on multimodal planning and reasoning \cite{huang2025thinkact,yang2025visionthink}.

\subsection{Video Understanding Datasets and Benchmarks}
The evolution of multimodal benchmarks has driven progress from visual-only QA to comprehensive audiovisual reasoning. 
Early datasets such as MSRVTT-QA \cite{xu2017msrvttqa}, and ActivityNet-QA \cite{yu2019activitynet} focus on text–video understanding, while recent ones including MovieChat \cite{song2024moviechat}, Video-MME \cite{fu2025videomme}, and MMBench-Video \cite{fang2024mmbench} extend evaluation to open-ended, multi-turn, and long-video reasoning. 
Audiovisual datasets such as AVQA \cite{yang2022avqa} and Music-AVQA \cite{Li2022Learning} further integrate sound cues for joint perception. 
Despite Despite broader domain coverage, recent resources still face challenges: WorldSense \cite{hong2025worldsense} requires costly manual annotation, Daily-Omni \cite{zhou2025dailyomni} suffers from semantic discontinuity due to fixed-duration segmentation, and IntentBench \cite{yang2025humanomniv2} offers limited novelty by aggregating prior datasets. 
Persistent gaps in event segmentation, filtering quality, multi-event reasoning, and long-video understanding motivate our benchmark for robust audiovisual reasoning.

\begin{figure*}[t]
  \centering
  \includegraphics[width=\textwidth]{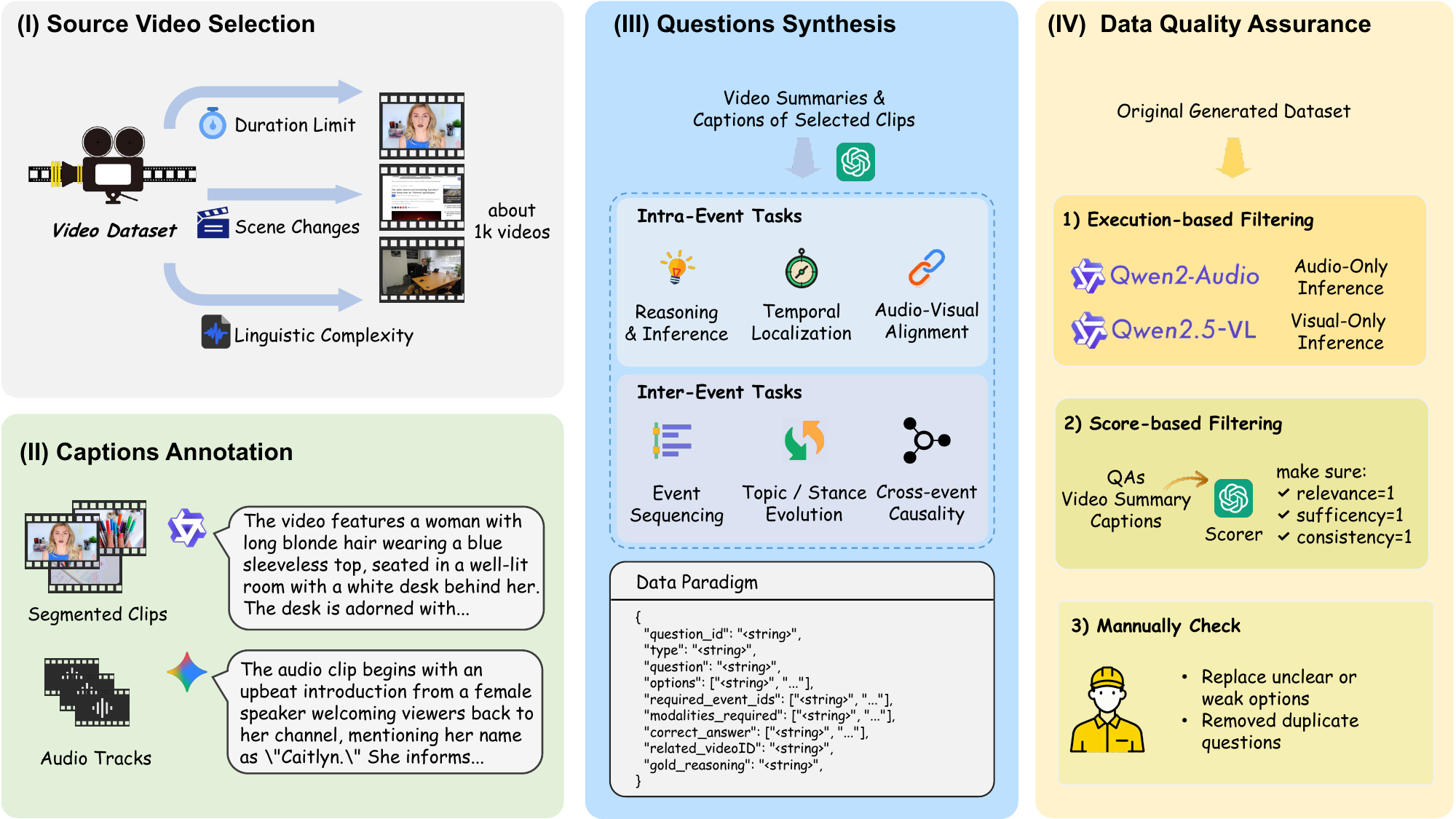}
  \caption{\textbf{Overview of the LongInsightBench construction workflow.} The pipeline begins with \textbf{video selection} from FineVideo, applying filters on duration, scene shifts, and content richness. Next, \textbf{automated annotation} integrates visual and audio descriptions via MLLMs. These annotations support \textbf{task scenario design and question generation}, spanning intra-event and inter-event reasoning tasks. Finally, a \textbf{quality assurance process} combines automatic filtering, scoring, and manual validation to ensure a high-quality QA set.}
  \label{fig:overview}
  \vspace{-0.1in}
\end{figure*}

\subsection{Multimodal Emotion Recognition and Intent Understanding}

Current AI research in Emotional Intelligence (EI) is comprehensive, covering practical applications like the empathetic dialogue model Emohaa, and systematic multi-modal assessments like HumanSense~\cite{qin2025humansense} and the psychology-based EmoBench~\cite{sabour-etal-2024-emobench}. Specialized benchmarks further contribute, with MEMO-Bench~\cite{zhou2024memobenchmultiplebenchmarktexttoimage} evaluating emotion in generated content and models like Qieemo~\cite{chen2025qieemospeechneedemotion} advancing foundational audio emotion recognition. Despite this wide coverage across modalities (text, audio, visual) and tasks (from perception to application), these works predominantly rely on static images, short dialogues, or brief video clips. This focus limits their ability to evaluate the dynamic, long-range evolution of emotional and social contexts, which require tracking subtle, non-contiguous events over extended periods.

\begin{table*}[t]
\centering
\renewcommand{\arraystretch}{1.2} 
\small
\resizebox{\textwidth}{!}{ 
\begin{tabular}{l|c|r|r|r|c|c|c|c|c}
\toprule
\textbf{Benchmarks} & \textbf{Mod.} & \textbf{\#Vids} & \textbf{Dur.(s)} & \textbf{\#QA} & \textbf{Anno.} & \textbf{Multi} & \textbf{Open} & \textbf{A-V Corr.} & \textbf{Emo\&Insight} \\
\midrule
MSRVTT-QA      & V   & 2,990  & 15.2   & 72,821 & A   & \xmark & \cmark & \xmark & \xmark \\
ActivityNet-QA & V   & 800    & 111.4  & 8,000  & M   & \xmark & \xmark & \xmark & \xmark \\
MVBench        & V   & 3,641  & 16.0   & 4,000  & A   & \xmark & \cmark & \xmark & \xmark \\
MovieChat      & V   & 130    & 500.0  & 1,950  & M   & \xmark & \cmark & \xmark & \xmark \\
Video-Bench    & V   & 5,917  & 56.0   & 17,036 & A\&M& \xmark & \cmark & \xmark & \xmark \\
EgoSchema      & V   & 5,063  & 180.0  & 5,063  & A\&M& \cmark & \cmark & \xmark & \xmark \\
Video-MME      & V   & 900    & 1017.9 & 2,700  & M   & \xmark & \cmark & \cmark & \xmark \\
MMBench-Video  & V   & 609    & 165.4  & 1,998  & M   & \cmark & \cmark & \xmark & \xmark \\
\midrule
AVQA           & A+V & 57,000 & 10     & 57,335 & M   & \xmark & \cmark & \cmark & \xmark \\
Music-AVQA     & A+V & 9,288  & 60     & 45,867 & M   & \xmark & \cmark & \cmark & \xmark \\
OmniBench      & A+I & -      & -      & 1,142  & M   & \cmark & \cmark & \xmark & \xmark \\
AV-Odyssey     & A+I & -      & -      & 4,555  & M   & \cmark & \cmark & \xmark & \xmark \\
LongVALE       & A+V & 8,400  & 235    & -      & A\&M& \cmark & \cmark & \cmark & \xmark \\
WorldSense     & A+V & 1,662  & 141.1  & 3,172  & M   & \cmark & \cmark & \cmark & \xmark \\
Daily-Omni     & A+V & 684    & 30--60 & 1,197  & A\&M& \cmark & \cmark & \cmark & \xmark \\
\midrule
\textbf{LongInsightBench} & A+V & 1,001  & 539.1 & 4,781  & A\&M & \cmark & \cmark & \cmark & \cmark \\
\bottomrule
\end{tabular}}
\caption{\textbf{Statistics of representative video QA benchmarks.} 
Mod. denotes modality. Dur.(s) is mean video duration in seconds. 
Anno. indicates automatic (A) or manual (M) annotations. 
Multi shows whether the dataset includes multiple question types. 
Open signifies coverage of diverse domains. 
A-V Corr. specifies if multimodal integration is required. 
Emo\&Insight highlights whether the benchmark focuses on recognizing human intentions, emotions, and language-driven elements.}
\label{tab:benchmarks}
\vspace{-0.1in}
\end{table*}

\section{LongInsightBench Dataset Construction}

To ensure LongInsightBench serves as a challenging and reliable benchmark for evaluating omni-modal understanding in long videos, we established a rigorous, multi-stage construction pipeline(See Figure~\ref{fig:overview}). This process involved careful video selection, stringent filtering based on multi-modal density, automated captioning, structured question generation across diverse tasks, and a comprehensive quality assurance protocol.

\subsection{Video Selection and Filtering}

\subsubsection{Video Source and Categories}
The video corpus for LongInsightBench is sourced from the publicly available FineVideo dataset \cite{Farré2024FineVideo}, with a strategic focus on content categories that naturally exhibit high linguistic complexity and diverse multi-modal interactions rather than mere object recognition, which is crucial for testing deep comprehension. 


\subsubsection{Filtering for Temporal Scope and Information Richness}
\label{sec:video_filter}
To ensure the selected videos present a significant temporal challenge and require deep, long-range reasoning, we applied two strict filtering criteria beyond the initial category selection.
\vspace{-0.2em}
\paragraph{Duration Constraint.} All selected videos must be longer than 7 minutes. This threshold ensures that models are challenged in maintaining contextual awareness and managing long-term temporal dependencies.
\vspace{-0.2em}
\paragraph{Criteria for Content Richness.} 
We established criteria for content richness across both visual and audio-textual modalities. 
For \textbf{visual dynamism}, we utilized \texttt{pyscenedetect} to identify scene cuts. Only videos exhibiting at least three distinct scene changes were retained, ensuring visual diversity and dynamic content. 
Regarding \textbf{linguistic complexity}, we transcribed the audio using WhisperX (with Whisper-medium model and Wav2Vec2.0-based alignment module) and removed non-English content. Furthermore, to verify the complexity of the argumentative structure, we employed GPT-4o for paragraph-level semantic segmentation, retaining only videos that demonstrated at least 4 distinct topic shifts throughout their duration. The details of semantic segmentation is listed in Appendix~\ref{sec:seg}.

\subsection{Automated Annotation of Captions}

To facilitate precise question generation and subsequent automated evaluation, we performed detailed multi-modal annotation on the filtered video corpus.

Based on the segmentation result introduced in Section \ref{sec:video_filter}, we generated specialized multi-modal captions for each clip. The visual captions are generated using Ovis2.5-9B, focusing on describing the actions, entities, and visual context within the clip, while the audio captions are annotated using Gemini2.0-Flash, focusing on summarizing the spoken content, identifying speaker sentiment, and describing music or background sounds.

\subsection{Task Scenarios and Question Generation}
LongInsightBench is designed to evaluate multi-modal reasoning abilities across two complementary dimensions: fine-grained perception within short temporal spans and holistic understanding over extended durations. To this end, we define six task scenarios, grouped into two categories: \textbf{Intra-event tasks}, targeting localized reasoning, and \textbf{Inter-event tasks}, targeting long-range reasoning across events.

While the questions adopt a multiple-choice format, the number of valid answers is dynamically determined by the LLM, in order to increase reasoning  diversity and complexity. 

\textbf{Intra-event Tasks.}
These tasks evaluate a model's ability to perceive and reason within a short temporal window for about one minutes, requiring multimodal alignment and local inference:
\begin{itemize}[itemsep=1pt, topsep=2pt, parsep=1pt, partopsep=0pt, leftmargin=1.5em]
    \item \textbf{Reasoning/Inference:} 
    Local causal or inferential reasoning based on immediately preceding or simultaneous multimodal cues.
    \item \textbf{Temporal Localization:} Identifying the precise timing of a specific event or action, using visual and auditory evidence.
    \item \textbf{Audio-Visual Alignment:} Matching spoken or textual content with corresponding visual cues to ensure cross-modal consistency.
\end{itemize}

\textbf{Inter-event Tasks.} 
These tasks assess long-range understanding, evaluating a model's ability to retain, integrate, and reason over information distributed across the video timeline, which often require combining multiple non-contiguous events.
\begin{itemize}[itemsep=1pt, topsep=2pt, parsep=1pt, partopsep=0pt, leftmargin=1.8em]
    \item \textbf{Timeline Reconstruction:} 
    Sequencing events across the video by linking distant audio–visual anchors.
    \item \textbf{Topic/Stance Evolution:} Tracking how specific themes, viewpoints, or arguments develop, shift, or evolve throughout the video.
    \item \textbf{Cross-event Causality:} Reasoning over long temporal gaps to uncover causal relationships connecting earlier triggers and later outcomes.
\end{itemize}

We employ GPT-4o to generate QA pairs with a "gold reasoning" for each task type using tailored prompt templates, as detailed in Appendix~\ref{sec:qa_prompt}.

\subsection{Rigorous Quality Assurance Pipeline}

To ensure that the generated questions truly required multi-modal, long-range reasoning. We implemented a three-step, semi-automated filtering pipeline.
\vspace{-0.5em}
\paragraph{Step 1: Execution-Based Filtering.}  
We used state-of-the-art single-modality models as solvers to identify and remove questions not requiring multimodal fusion. Questions solvable by \textbf{Qwen2.5-VL-7B-Instruct} (vision-text only) or \textbf{Qwen2-Audio-7B-Instruct} (audio-text only) were discarded.
\vspace{-0.5em}
\paragraph{Step 2: Score-Based Filtering.}  

Each remaining QA pair was automatically reviewed by \textbf{GPT-4o} along three scoring dimensions:  
(1) \textbf{Relevance} — whether the question truly depends on the video content rather than general knowledge or common sense;  
(2) \textbf{Sufficiency} — whether the provided multi-modal inputs contain enough evidence for a correct answer; and  
(3) \textbf{Consistency} — whether the answer is factually correct and logically aligned with the ground truth from the video segment.  
Each criterion was rated between 0 and 1, and only QA pairs achieving high scores across all three dimensions were retained. Details of the scoring prompt and criteria are provided in Appendix~\ref{sec:sco_prompt}.



\vspace{-0.3em}
\paragraph{Step 3: Manual Inspection.}  
A random sample of the filtered QA pairs was checked by human annotators. During this review, the annotators replaced unclear or weak answer options, removed questions that were too similar to each other, and adjusted the difficulty level where needed.

\begin{table}[h]
  \centering
  \begin{tabular}{lcc}
    \toprule
    \textbf{Task Type} & \textbf{Initial} & \textbf{Filtered} \\
    \midrule
    Intra-event Reasoning & 2002 & 855 \\
    Temporal Localization & 2002 & 857 \\
    Audio-Visual Alignment & 2002 & 1307 \\
    Timeline Reconstruction & 2002 & 506 \\
    Topic/Stance Evolution & 626 & 165 \\
    Cross-event Causality & 2002 & 1091 \\
    \midrule
    Total & 10636 & 4781 \\
    \bottomrule
  \end{tabular}
  \caption{\textbf{Final dataset statistics} after multi-stage filtering.}
  \label{tab:qa_stats}
  \vspace{-0.1in}
\end{table}

\begin{table*}[t]
\centering
\renewcommand{\arraystretch}{1.0} 
\resizebox{\textwidth}{!}{
\begin{tabular}{
    l!{\vrule width 0.4pt\hspace{0.1em}\vrule width 0.4pt}  
    ccc!{\vrule width 0.4pt}  
    ccc!{\vrule width 0.4pt}  
    c
}
\toprule
\multirow{2}{*}{\textbf{Model}} & \multicolumn{3}{c!{\vrule width 0.4pt}}{\textbf{Intra-event}} & \multicolumn{3}{c!{\vrule width 0.4pt}}{\textbf{Inter-event}} & \multirow{2}{*}{\textbf{Overall}} \\
\cmidrule(lr){2-7}
& \textbf{IE-Rea} & \textbf{T-Loc} & \textbf{AV-Align} & \textbf{T-Recon} & \textbf{Topic Evo\&Sum} & \textbf{CE-Caus}\\
\midrule
Qwen2.5-Omni-7B    & 61.75 & 20.77 & 49.66 & 52.57 & \cellcolor{red!25}\textbf{89.70} & \cellcolor{red!25}\textbf{49.95} & 48.40 \\
VideoLLama3        & 59.30 & \phantom{0}9.92 & 56.69 & 45.85 & 63.03 & 19.52 & 39.36 \\
VideoLLama2        & 38.36 & 13.89 & 37.03 & 47.04 & 46.67 & \phantom{0}9.35 & 28.19 \\
Ola-7B             & \cellcolor{red!10}\textbf{70.18} & \cellcolor{red!10}\textbf{27.19} & \cellcolor{red!10}\textbf{63.20} & \cellcolor{red!10}\textbf{63.44} & 63.03 & 39.14 & \cellcolor{red!10}\textbf{52.52} \\
Unified-IO-2 L     & 27.25 & \cellcolor{prompt_blue!75}\textbf{\phantom{0}2.10} & \cellcolor{prompt_blue!75}\textbf{24.10} & 25.49 & 36.36 & \phantom{0}8.80 & \cellcolor{prompt_blue!75}\textbf{17.80} \\
Unified-IO-2 XL    & \cellcolor{prompt_blue!75}\textbf{18.25} & \phantom{0}6.07 & 39.40 & \cellcolor{prompt_blue!75}\textbf{22.92} & \cellcolor{prompt_blue!75}\textbf{29.09} & \phantom{0}2.47 & 19.12 \\
Unified-IO-2 XXL   & 36.84 & 17.85 & 54.55 & 32.81 & 50.30 & \cellcolor{prompt_blue!75}\textbf{\phantom{0}1.10} & 30.16 \\
Gemini2.5-Flash    & \cellcolor{red!25}\textbf{89.01} & \cellcolor{red!25}\textbf{38.86} & \cellcolor{red!25}\textbf{76.05} & \cellcolor{red!25}\textbf{86.76} & \cellcolor{red!10}\textbf{84.24} & \cellcolor{red!10}\textbf{41.25} & \cellcolor{red!25}\textbf{65.17} \\
\bottomrule
\end{tabular}
}
\caption{\textbf{Average accuracy (\%) of various OLMs across different tasks.} Abbreviations: IE-Rea (Intra-event Reasoning), T-Loc (Multimedia Temporal Localization), AV-Align (Audio-Visual Alignment), T-Recon (Timeline Reconstruction), Topic Evo\&Sum (Topic/Stance Evolution Summarization), CE-Caus (Cross-event Causality).}
\label{tab:omni_res}
\vspace{-0.1in}
\end{table*}

\subsection{LongInsightBench Statistics}

As summarized in Table~\ref{tab:benchmarks}, our proposed benchmark consists of 1001 videos and 4781 high-quality QA pairs after a rigorous three-step filtering pipeline. The average video duration is 539 seconds, which is substantially longer than existing audio-visual understanding benchmarks. 
The approximately 1,000 selected videos are divided into three main categories: the lecture category, which includes 517 videos across 8 subcategories; the interview category, comprising 258 videos across 4 subcategories; and the Vlogs/Film Trailers category, with 230 videos spanning 4 subcategories.
Detailed descriptions of the video categories are provided in Table~\ref{tab:video_source}. The distribution of the collected videos across all subcategories is visualized in Figure~\ref{fig:video_distributions}.
The task-type distribution in Table~\ref{tab:qa_stats} demonstrates that the benchmark covers a wide spectrum of reasoning scenarios. From localized Intra-event Reasoning and Audio-Visual Alignment to global-level Cross-event Causality and Timeline Reconstruction, the benchmark requires models to handle both fine-grained grounding and complex long-horizon dependencies. 
Such diversity makes the evaluation more comprehensive.

\section{Experiments and Analysis}

\subsection{Settings}
We evaluate four categories of multimodal large language models (MLLMs): 
(1) \textbf{OLMs}, including open-source contenders like VideoLLaMA 2 \cite{damonlpsg2024videollama2}, VideoLLaMA 3 \cite{damonlpsg2025videollama3}, Unified-IO-2 \cite{lu2023uio2}, Qwen2.5-Omni \cite{Qwen2.5-Omni}, Ola \cite{liu2025ola}, and the proprietary model Gemini2.5-Flash \cite{comanici2025gemini25}; 
(2) \textbf{VLMs}, represented by the open-source Ovis2.5 \cite{lu2025ovis25technicalreport} and proprietary GPT-4o \cite{openai2024gpt4o}, Gemini2.5-Flash; 
(3) \textbf{ALMs}, such as Gemini2.5-Flash; and (4) \textbf{LLMs}, represented by GPT-4o.
All evaluations follow each model’s official inference pipeline and pre-processing configurations. 
To ensure consistency across model types, we set the number of video frames to $64$ for open-source models, while proprietary models are evaluated through official APIs using default multimodal input settings.

Our experiments consist of three parts. 
First, we evaluate a wide range of OLMs on the full benchmark to compare their abilities in perceiving multimodal linguistic cues in long videos. 
Second, using LongInsightBench, we vary modality inputs: (i) VLMs with videos and audio captions, (ii) ALMs with audios and visual captions, and (iii) LLMs with both visual and audio captions, to examine whether integrating multiple modalities introduces losses or biases associated with specific modalities. 
Third, we conduct ablation studies varying the number of sampled video frames to assess open-source OLMs under different frame sampling rates. 
Performance in all experiments is measured by accuracy, with a prediction considered correct only if all options are exactly selected, ensuring a strict assessment of the model’s multimodal understanding.

To estimate the total computational budget, we consider the local model deployed on our own servers, utilizing 16 NVIDIA A800-SXM4-40GB GPUs, which require approximately 250 GPU Hours for the entire experiment, with no additional costs. The server is equipped with CUDA version 13.0 and NVIDIA driver version 580.65.06, supporting the necessary computational load. For GPT-4o and Gemini2.5-Flash models, cloud services are used, resulting in a total estimated cost of \$200 for the API calling.


\subsection{Main Results}

The experimental results presented in Table 1 provide a comprehensive evaluation of various Omni-modal Language Models (OLMs) across the six different task scenarios defined in LongInsightBench. The analysis clearly shows the current performance gap between proprietary, large-scale models and their open-source counterparts, while also highlighting specific areas of weakness that are common across all models, particularly in long-range temporal and causal reasoning.

\paragraph{Performance Ceiling and Model Hierarchy}
The results clearly show that \textbf{Gemini2.5-Flash} is the best performer on this benchmark, with an overall accuracy of \textbf{$0.6517$}. This closed-source model scored the highest or second-highest in five out of the six individual task categories, demonstrating strong performance and reliability across both localized and long-range tasks.

Among the open-source models,\textbf{ Ola-7B} stands out as the best competitor, achieving the second-highest overall score (\textbf{$0.5252$}) and ranking second in all three Intra-event tasks (IE-Rea, T-Loc, AV-Align) and Timeline Reconstruction (T-Recon). This suggests that Ola-7B has a solid base for multi-modal understanding and localized inference.

On the other hand, the \textbf{Unified-IO-2} models generally performed poorly, placing in the lowest performance groups. Unified-IO-2 L recorded the lowest overall accuracy (\textbf{$0.1780$}), and its variants performed poorly in several key areas, such as Intra-event Reasoning (IE-Rea), Temporal Localization (T-Loc), and Cross-event Causality (CE-Caus), suggesting that these models lack the needed multi-modal alignment and long-context processing capabilities for this benchmark.

\begin{figure}[t]
     \centering
     \includegraphics[width=1\linewidth]{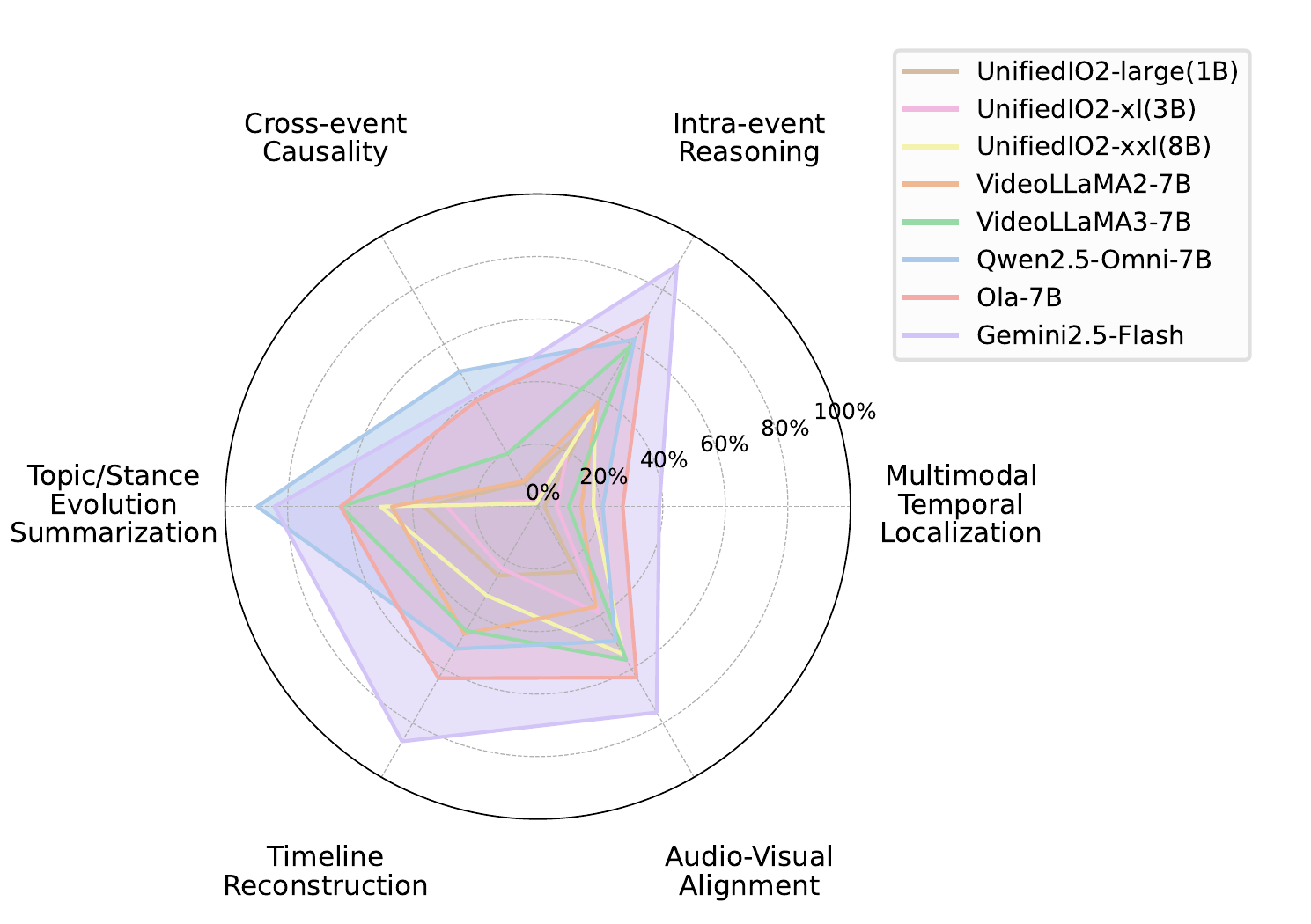}
     \caption{\textbf{Fine-grained performance across task categories.} Different OLMs’ accuracies are shown over six question types, highlighting each model’s strengths and weaknesses across categories.}
    \label{fig:category_result}
    \vspace{-0.2in}
\end{figure}

\paragraph{Analysis of Task Categories}
Figure~\ref{fig:category_result} visualizes the model performance hierarchy ,showing clear differences between tasks in difficulty, effectively testing the limits of current OLMs.

Performance on the Intra-event tasks (IE-Rea, T-Loc, AV-Align) is generally better than on the Inter-event tasks, reflecting the relative ease of localized processing. However, Temporal Localization (T-Loc) proved to be a major challenge for all models. Even the top performer, Gemini2.5-Flash, only scored \textbf{$0.3886$}, and the scores were much lower for other models (e.g., Unified-IO-2 L scored $0.0210$). This shows that accurately pinpointing event timing using multi-modal data remains a major challenge, requiring further improvements in fine-grained temporal modeling. Audio-Visual Alignment (AV-Align) saw strong performance from Gemini (\textbf{$0.7605$}), confirming its ability to align audio and visual data effectively.

The Inter-event tasks, which require long-term memory and synthesis, showed the most variation. The task Topic/Stance Evolution Summarization seemed relatively easier, with Qwen2.5-Omni-7B unexpectedly achieving the highest score (\textbf{$0.8970$}) and Gemini close behind ($0.8424$). This indicates that many models are good at tracking and summarizing the main theme or narrative flow across the video.

In contrast, Cross-event Causality (CE-Caus) proved to be the most difficult long-range task. This task involves identifying cause-and-effect relationships across different parts of the video, and performance was generally low. Gemini achieved the highest score at \textbf{$0.4125$}, while several models, including Unified-IO-2 XXL, scored near chance level (\textbf{$0.0110$}). This highlights the current limitations of OLMs in maintaining and reasoning over complex causal links within long videos.

In summary, LongInsightBench effectively shows the differences in model capabilities, demonstrating the clear advantage of large proprietary models like Gemini2.5-Flash. The benchmark also shows that while models are improving in tracking narrative themes (Topic Evo\&Sum), they still face significant challenges in precise temporal localization (T-Loc) and complex, long-range causal reasoning (CE-Caus).

\begin{table}[h!]
\centering
\resizebox{\linewidth}{!}{
\begin{tabular}{l|ccc}
\toprule
\textbf{Models} & \textbf{Intra-event} & \textbf{Inter-event} & \textbf{Overall} \\
\midrule
\addlinespace[-0.2pt]
\multicolumn{4}{c}{\cellcolor{model_type}\textbf{VLM + audio captions}} \\
\addlinespace[-1.5pt]
\midrule
Ovis2.5-9B & 54.46 & 60.11 & 56.55 \\
GPT-4o & 64.03 & 63.48 & 63.83 \\
Gemini2.5-Flash & 72.28 & 60.11 & 67.78 \\
\midrule
\addlinespace[-0.2pt]
\multicolumn{4}{c}{\cellcolor{model_type}\textbf{ALM + visual captions}} \\
\addlinespace[-1.5pt]
\midrule
Gemini2.5-Flash & 72.94 & 64.04 & \textbf{69.65} \\
\midrule
\addlinespace[-0.2pt]
\multicolumn{4}{c}{\cellcolor{model_type}\textbf{LLM + both captions}} \\
\addlinespace[-1.5pt]
\midrule
GPT-4o & 67.99 & 53.93 & 62.79 \\
\midrule
\addlinespace[-0.2pt]
\multicolumn{4}{c}{\cellcolor{model_type}\textbf{SOTA OLM}} \\
\addlinespace[-1.5pt]
\midrule
Gemini2.5-Falsh & 69.31 & 58.43 & 65.17 \\
Ola-7B & 59.08 & 49.44 & 52.52 \\
\bottomrule
\end{tabular}
}
\caption{\textbf{Performance comparison} (accuracy, \%) of VLMs(with audio captions), ALMs(with visual captions), LLMs(with both captions) and SOTA OLMs.}
\label{tab:modal}
\vspace{-0.1in}
\end{table}

\subsection{Singal-modal Models v.s. Omni-modal Models}

The comparative analysis presented in Table~\ref{tab:modal} contrasts the performance of dedicated Signal-modal Models (VLM, ALM) and LLMs, which replace one or two of the input modalities with textual descriptions, against true Omni-modal Models (OLM), revealing critical insights into the current state of multi-modal fusion.

\paragraph{The Paradox of Omni-modal Fusion}
The most striking observation is the performance paradox exhibited by the state-of-the-art model, Gemini2.5-Flash. When processing both raw visual and raw audio data simultaneously (OLM: $0.6517$), its performance is significantly lower than when one modality is replaced by a high-quality textual description (VLM: $0.6778$, ALM: $\mathbf{0.6965}$). The highest overall score is achieved by the Audio-Language Model (ALM) configuration, suggesting that the model excels when the visual information is provided in a distilled, textually abstracted format.

This phenomenon strongly suggests that current OLM fusion mechanisms suffer from a \textit{fusion deficit}, where the process of integrating two raw modalities (pixels and waveforms) introduces either \textit{information loss} or \textit{processing bias}.

\begin{table}[t!]
\centering
\resizebox{\linewidth}{!}{
\begin{tabular}{lccc}
\toprule
\multirow{2}{*}{\textbf{Model}} & \multicolumn{3}{c}{\textbf{Input Frame Number}} \\
\cmidrule(lr){2-4}
& \textbf{32} & \textbf{64} & \textbf{128} \\
\midrule
VideoLLama3 & 39.29 & 41.37 & 43.45 \\
Ola-7B & 53.64 & 55.30 & 56.96 \\
Unified-IO-2 XXL & 30.98 & 31.19 & 31.39 \\
Qwen2.5-Omni-7B & 51.35 & 51.98 & 51.98 \\
\bottomrule
\end{tabular}
}
\caption{Overall Accuracy (\%) with \textbf{Different Input Frame Numbers}.}
\label{tab:frame}
\vspace{-0.1in}
\end{table}

\paragraph{Explaining the Superiority of Textual Proxies}



The superior performance of VLM and ALM configurations, which use textual descriptions for one modality, stems from three factors.

First, textual descriptions (captions/summaries) act as effective, pre-processed proxies, filtering noise and redundancy from raw streams. This allows the model to bypass complex, error-prone low-level feature extraction and alignment for that modality.
Second, receiving one modality as text enables the model to dedicate its full capacity (e.g., attention) to robustly aligning the remaining raw modality with the pre-parsed text. This focused processing leads to a more accurate overall understanding.
Third, these models are fundamentally rooted in Large Language Models (LLMs), which operate optimally with high-quality textual input. This is supported by GPT-4o's competitive score ($0.6279$) using purely textual input. Ultimately, when the raw fusion mechanism is imperfect, the quality of textual representation can often outweigh the benefit of processing raw multi-modal data.

\subsection{The Effect of Video Frame Sampling Rate}

The ablation study presented in Table~\ref{tab:frame} examines the impact of visual information density by varying the number of sampled input video frames (32, 64, and 128) on open-source OLM performance. The results generally confirm that increased frame sampling leads to improved overall accuracy, though the degree of benefit is highly model-dependent.

\textbf{Models Showing Strong Context Utilization} (Ola-7B and VideoLLama3):
Ola-7B and VideoLLama3 demonstrated a clear and consistent positive correlation between frame count and accuracy. Ola-7B, the top open-source model, saw its performance steadily increase from $0.5364$ (32 frames) to $0.5696$ (128 frames). VideoLLama3 exhibited a similar robust trend ($0.3929$ to $0.4345$). This suggests that these architectures possess effective temporal integration mechanisms capable of successfully leveraging denser visual evidence to enhance long-context reasoning.

\textbf{Models Showing Saturation or Insensitivity} (Qwen2.5-Omni-7B and UnifiedIO 2-XXL):
In contrast, Qwen2.5-Omni-7B and UnifiedIO 2-XXL showed limited gains, indicating potential bottlenecks in their visual processing capacity. Qwen2.5-Omni-7B quickly reached a saturation point, improving slightly from 32 frames ($0.5135$) to 64 frames ($0.5198$) but showing no further improvement at 128 frames. UnifiedIO 2-XXL displayed extreme insensitivity, with its accuracy remaining nearly static across all tested frame counts ($0.3098$ to $0.3139$). This suggests that these models either struggle to integrate non-redundant information from denser sampling or that their internal visual context windows are saturated quickly, nullifying the benefit of increased frame input.











\section{Conclusion}
\label{sec:bibtex}

In conclusion, this paper introduces LongInsightBench, a pioneering benchmark for long-video omni-modal understanding, featuring a challenging dataset of 1,000 information-dense videos. This benchmark serves as a realistic testbed for next-generation OLMs, focusing on content rich in language-embedded cues such as viewpoint, sentiment, and action. Our experimental results demonstrate OLMs still face challenges in tasks like temporal localization and long-range causal reasoning. Additionally, extended experiment suggests that current omni-modal fusion mechanisms may suffer from a fusion deficit. The ablation study further reveals that frame sampling improves model accuracy, though the benefits vary across models. 

\section*{Limitations}

The development of LongInsightBench involved significant operational costs due to the reliance on proprietary models (GPT-4o and Gemini2.0-Flash) for high-fidelity filtering, QA generation, and rigorous quality assurance. This high API calling expense restricts the pace and scale at which we can expand the dataset, despite the need for dense, multi-modal content. This financial constraint is a major limitation on scalability. Future work will focus on developing more cost-efficient, open-source or human-in-the-loop pipelines to mitigate this expense and facilitate larger-scale expansion.



\section*{Ethics Statement}

We have carefully curated the video data used in LongInsightBench to ensure the exclusion of dangerous, discriminatory, or unhealthy content. Furthermore, we strictly adhere to all terms and conditions mandated by the source dataset, FineVideo. To respect the rights and privacy of the original video creators, we do not host the raw FineVideo content. Instead, we only release the synthesized data (questions, answers, and annotations) derived from the videos, maintaining responsible data usage within reasonable limits.



\bibliography{custom}

\appendix

\section{Video Source Details}
\begin{figure*}[t]
  \centering
  \includegraphics[width=\textwidth]{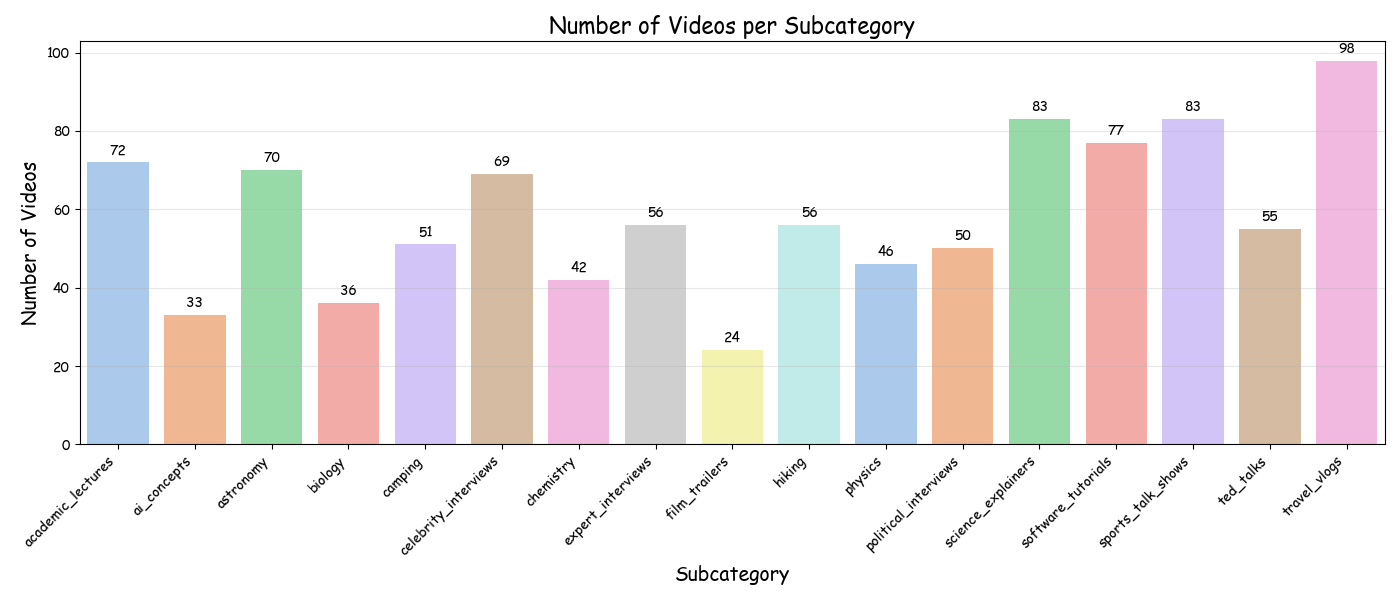}
  \captionsetup{width=0.95\textwidth}
  \caption{The \textbf{distribution} of videos in LongInsightBench across all subcategories.}
  \label{fig:video_distributions}
\end{figure*}

Table~\ref{tab:video_source} shows the details of all categories of video source. Figute~\ref{fig:video_distributions} visualizes the distribution of videos in our dataset across all subcategories.

\begin{table*}[h]
  \centering
  \begin{tabularx}{\textwidth}{lccX} 
    \toprule
    \textbf{Category} & \textbf{Subcategories} & \textbf{Count} & \textbf{Description} \\
    \midrule
    Lectures & 8 & 514 & Academic talks and tutorials covering diverse topics including AI, astronomy, biology, chemistry, science explanations, software tutorials, and TED talks. High focus on spoken content and visual aids. \\
    Interviews & 4 & 258 & Dialogues and interviews including celebrity, expert, and political interviews and sports talk shows. Rich in viewpoint tracking, sentiment changes, and nuanced discussion. \\
    Vlogs / Film Trailers & 4 & 229 & Narrative-driven content including camping, hiking, travel vlogs and film trailers. Emphasizes temporal coherence, action-language alignment, and multi-modal dynamics. \\
    \midrule
    Total & -- & 1001 & \\
    \bottomrule
  \end{tabularx}
  \caption{\textbf{Video source categories} from FineVideo used in LongInsightBench, updated with precise subcategory counts and descriptions.}
  \label{tab:video_source}
\end{table*}

\section{Implementation Details}
\subsection{Details of Semantic Segmentation}
\label{sec:seg}
To prevent the model from modifying the original transcript due to hallucinations, we design a two-stage semantic segmentation procedure. 
Both stages are performed using GPT-4o to ensure consistency in linguistic style and reasoning.
The prompts used in each stage are provided in \ref{sec:seg_prompt}.

In the first stage, the model outlines the overall thematic structure of the transcript.
Given the full transcript of the video’s speech, it is prompted to estimate the number of distinct semantic topics and to generate a tentative title for each.
This serves as a preparatory step for segmentation. 
Since the transcripts are often lengthy, identifying the expected number and focus of segments in advance helps reduce the model’s cognitive load in the subsequent stage.

In the second stage, the model determines the precise semantic boundaries between segments.
A second-stage prompt instructs the model to locate transition points between topics without modifying the original text.
To indicate these transitions, the model outputs a few words surrounding each boundary, from which the full text segments are later extracted using regular expressions.
The prompt explicitly requires that boundaries be placed between sentences. 
However, if a predicted boundary still falls within a sentence, the entire sentence is assigned to the following segment, while the preceding sentence serves as the closure of the previous one.

We adopt this boundary-marking strategy rather than asking the model to directly output segmented text, in order to avoid discontinuities and hallucinations.
LLMs may inadvertently add, omit, or alter portions of the transcript especially when handling long passages, resulting in inconsistencies or incomplete context for subsequent audio-captioning tasks.

\subsection{Prompts for visual/audio caption}
\label{sec:av_prompt}
Prompt in Figure \ref{fig:av-prompt-1} is used for creating visual caption by Ovis2.5-9B, and Prompt in Figure \ref{fig:av-prompt-2} is used for creating audio caption by Gemini2.5-flash.

\begin{figure*}[h!]
\begin{tcolorbox}[
    colback=prompt_yellow!5!white,
    colframe=frame3,
    title=VISUAL CAPTION PROMPT,
    fonttitle=\bfseries,
    colbacktitle=prompt_yellow!45!white,
    coltitle=frame3!90!white,
    boxrule=1.5pt,
    arc=5pt,
    boxsep=5pt,
    left=12pt,
    right=12pt,
    top=12pt,
    bottom=12pt
]
You are an expert video describer.\\

Provide a detailed description of the given video segment as a single concise paragraph.\\

\textbf{Focus on:}\\
- People: their actions, gestures, clothing, and facial expressions (use distinguishing features to tell individuals apart)\\
- Objects and text: describe visible objects and any on-screen text (state the text in its original language, give an English translation in parentheses, and explain its contextual meaning)\\
- Environment: the setting, background details, and atmosphere\\
- Visual changes: transitions, movements, or notable differences between frames\\

\textbf{Guidelines:}\\
- Describe the sequence of frames as a continuous narrative, not isolated snapshots\\
- Emphasize how the scene evolves over time\\
- Avoid speculation beyond what is visually shown\\
\end{tcolorbox}
\caption{Prompt used for providing \textbf{visual captions} for each video.}
\label{fig:av-prompt-1}
\vspace{-0.1in}
\end{figure*}

\begin{figure*}[h!]
\begin{tcolorbox}[
    colback=prompt_yellow!5!white,
    colframe=frame3,
    title=AUDIO CAPTION PROMPT,
    fonttitle=\bfseries,
    colbacktitle=prompt_yellow!45!white,
    coltitle=frame3!90!white,
    boxrule=1.5pt,
    arc=5pt,
    boxsep=5pt,
    left=12pt,
    right=12pt,
    top=12pt,
    bottom=12pt
]
You are an expert audio describer.\\

Provide a chronological description of the audio clip without mentioning timestamps.\\

\textbf{For speech, include:}\\
- Content (quote short phrases verbatim, summarize longer parts)\\
- Speaking tone\\
- Number of speakers, distinguishable by gender, speaking style, or any other perceivable audio features\\
\textbf{For music, include:}\\
- Genre or style\\
- Mood or tone\\
- Main instruments or notable features\\
\textbf{For background or ambient sounds, include:}\\
- Sound characteristics (volume, rhythm, consistency, etc.)\\
- Environmental cues or setting inferred from these sounds\\
\textbf{For other sounds, include:}\\
- Type of sound and how it contributes to the scene\\

\textbf{Guidelines:}\\
- Avoid guessing when uncertain \\
- Write the description as a single concise paragraph, highlighting transitions between different sounds
\end{tcolorbox}
\caption{Prompt used for providing \textbf{audio captions} for each video.}
\label{fig:av-prompt-2}
\vspace{-0.1in}
\end{figure*}

\subsection{Prompts for Semantic Segmentation}
\label{sec:seg_prompt}
Figure \ref{fig:seg-prompt-1} and \ref{fig:seg-prompt-2} illustrate the prompt design for semantic segmentation.
Particularly, the SEGMENT\_COUNT\_PROMPT shown in 
Figure \ref{fig:seg-prompt-1} is used in the first stage of segmentation, while the BOUNDARY\_DETECTION\_PROMPT shown in 
Figure \ref{fig:seg-prompt-2} is used in the second stage.

\begin{figure*}[h!]
\begin{tcolorbox}[
    colback=prompt_green!5!white,
    colframe=frame1,
    title=SEGMENT COUNT PROMPT,
    fonttitle=\bfseries,
    colbacktitle=prompt_green!50!white,
    coltitle=frame1,
    boxrule=1.5pt,
    arc=5pt,
    boxsep=5pt,
    left=12pt,
    right=12pt,
    top=12pt,
    bottom=12pt
]

You are an expert in transcript chunking and topic boundary detection for long videos.\\

Given a piece of text transcribed from the audio of a video, your task is to:\\
1. Identify how many distinct semantic chunks it contains. \\
2. For each chunk, provide a short title (a few words) summarizing its main theme or idea.\\

\textbf{Guidelines:}\\
- Each chunk should correspond to a coherent theme, explanation, or dialogue unit.  \\
- Avoid making chunks too short or too long. \\ 
- The goal of chunking is to create useful and self-contained units of text for downstream tasks such as captioning and retrieval, not to detect strict topic shifts.  \\
- The short titles should be concise, descriptive, and capture the main semantic focus of the chunk.  \\

\textbf{Output Format} (strictly follow this structure):\\
Chunk count: <integer>\\
Titles:\\
1. <short title for chunk 1>\\
2. <short title for chunk 2>\\
...\\
N. <short title for chunk N>\\

Now, analyze the following text:\\
\{text\}
\end{tcolorbox}
\caption{\textbf{Prompt used in the first stage of semantic segmentation.} In this stage, the model is asked to identify the number of topics of the given transcript.}
\label{fig:seg-prompt-1}
\vspace{-0.1in}
\end{figure*}

\begin{figure*}[h!]
\begin{tcolorbox}[
    colback=prompt_green!5!white,
    colframe=frame1,
    title=BOUNDARY DETECTION PROMPT,
    fonttitle=\bfseries,
    colbacktitle=prompt_green!50!white,
    coltitle=frame1,
    boxrule=1.5pt,
    arc=5pt,
    boxsep=5pt,
    left=12pt,
    right=12pt,
    top=12pt,
    bottom=12pt
]
You are an expert in transcript segmentation for long videos. \\

\textbf{Your task:} \\
1. Identify EXACTLY \{boundary\_count\} semantic boundaries in the transcript, based on the \{topic\_count\} chunks and their titles. \\
2. Each boundary MUST be represented as: \\
    <last few words of previous sentence>[BORDER]<first few words of next sentence> \\

\textbf{Guidelines:} \\
- You must output ONLY one boundary per line. No explanations, no numbering, no extra words. \\
- The words on the left and right of [BORDER] MUST appear \textbf{exactly as in the original transcript}, with no paraphrasing. \\
- The left part must be the END of a sentence. The right part must be the START of the next sentence. \\
- Boundaries must align with the given semantic titles. \\
- Do not add or skip boundaries. The number of output lines MUST equal \{boundary\_count\}. \\

\textbf{Output Format} (strict): \\
<previous sentence ending>[BORDER]<next sentence beginning> \\
(repeated \{boundary\_count\} times, one per line) \\

Now process the following transcript: \\

Transcript: \\
\{text\} \\

Topic count: \{topic\_count\} \\
Chunk titles: \\
\{titles\} \\

Output: \\
\end{tcolorbox}
\caption{\textbf{Prompt used in the second stage of semantic segmentation.} In this stage, the model is asked to output segment borders and several surrounding words based on the given topics and titles.}
\label{fig:seg-prompt-2}
\vspace{-0.1in}
\end{figure*}

\subsection{Details of QA Construction}
\label{sec:qa_prompt}
To ensure the generated questions require deep contextual understanding—especially for Inter-event tasks—the input to the LLM included a high-level summary of the video (borrowed from the FineVideo dataset). This summary is necessary to provide the LLM with the global narrative context and thematic structure of the video, preventing the generation of questions based solely on localized, isolated events. Along with this summary, we provided randomly sampled event IDs (typically 2–3) and their corresponding visual and audio captions. The LLM was instructed to output a JSON-formatted selection question. The choice between single-choice and multi-choice was determined dynamically based on the complexity inherent in the task scenario.
Prompt in Figure \ref{fig:qa-prompt-0} is the system prompt used by GPT-4o in QA construction stage, while prompts in Figures \ref{fig:qa-prompt-1}, \ref{fig:qa-prompt-2}, \ref{fig:qa-prompt-3}, \ref{fig:qa-prompt-4}, \ref{fig:qa-prompt-5} and \ref{fig:qa-prompt-6} are the user prompts specified to generate QAs of different task types.

\begin{figure*}[h!]
\begin{tcolorbox}[
    colback=prompt_blue!5!white,
    colframe=frame2,
    title=SYSTEM PROMPT,
    fonttitle=\bfseries,
    colbacktitle=prompt_blue!45!white,
    coltitle=frame2!80!white,
    boxrule=1.5pt,
    arc=5pt,
    boxsep=5pt,
    left=12pt,
    right=12pt,
    top=12pt,
    bottom=12pt
]

You are a multimodal question generator specializing in video understanding. \\
Your task is to create high-quality multiple-choice questions (MCQs) for video understanding. \\

You are restricted to using ONLY the provided event list (visual\_caption, audio\_caption, timestamps). \\
Do not use external knowledge, hallucinated facts, or information not present in the events. \\
Each generated question must strictly follow the JSON schema below. \\

\end{tcolorbox}
\caption{\textbf{System prompt} used in QA construction stage.}
\label{fig:qa-prompt-0}
\vspace{-0.1in}
\end{figure*}

\begin{figure*}[h!]
\begin{tcolorbox}[
    colback=prompt_blue!5!white,
    colframe=frame2,
    title=INTRA EVENT REASONING USER PROMPT,
    fonttitle=\bfseries,
    colbacktitle=prompt_blue!45!white,
    coltitle=frame2!80!white,
    boxrule=1.5pt,
    arc=5pt,
    boxsep=5pt,
    left=12pt,
    right=12pt,
    top=12pt,
    bottom=12pt
]

video\_id: \{video\_id\}\\
summary: \{summary\}\\
events: \{events\_str\}\\

\textbf{Task requirements:}\\
    \textbf{1) Generate N=2 **Intra-event Reasoning** questions.} Each question MUST:\\
    - Focus on a single event, querying **causation or conclusions within its timestamp range** (e.g., "Why X happened / How Y was achieved / What Z signifies between [start\_time] and [end\_time]?").\\
    - Use **exactly 1 event\_id**, referring only to its content and **timestamps**.\\
    - Require BOTH visual and audio evidence from that event; single modality is INSUFFICIENT.\\
    - Offer plausible options **strictly based on the specific event's details**.\\
    - Be "single\_choice\_question" or "multiple\_choice\_question".\\

    \textbf{2) For answer options:}\\
    - Provide exactly 4 options: A, B, C, D.\\
    - Distractors must be consistent with event details but incorrect.\\
    - For "single\_choice\_question": exactly one correct option.\\
    - For "multiple\_choice\_question": at least two correct options.\\

    \textbf{3) For explanations:}\\
    - Justify the correct answer(s) by synthesizing information **as if you were observing the video directly** and field the `"gold\_reasoning"`.\\
    - Reasoning must detail inference steps, **explicitly referencing the single event\_id and both visual + audio evidence.**\\
\end{tcolorbox}
\caption{Prompt used in QA construction in the \textbf{Intra-event Reasoning} task.}
\label{fig:qa-prompt-1}
\vspace{-0.1in}
\end{figure*}

\begin{figure*}[h!]
\begin{tcolorbox}[
    colback=prompt_blue!5!white,
    colframe=frame2,
    title=MULTIMODAL TEMPORAL LOCALIZATION USER PROMPT,
    fonttitle=\bfseries,
    colbacktitle=prompt_blue!45!white,
    coltitle=frame2!80!white,
    boxrule=1.5pt,
    arc=5pt,
    boxsep=5pt,
    left=12pt,
    right=12pt,
    top=12pt,
    bottom=12pt
]

video\_id: \{video\_id\}\\
summary: \{summary\}\\
events: \{events\_str\}\\

\textbf{Task requirements:}\\
    \textbf{1) Generate N=2 **Multimodal Temporal Localization** questions.} Each question MUST:\\
    - Focus on localizing a specific event, which is defined by the **simultaneous occurrence or strong correlation of a distinct visual action/cue AND associated audio information (e.g., speech content, specific sounds)**.\\
    - Ask for the exact time segment(s).\\
    - Use **exactly 1 event\_id**. The question should provide enough detail from both visual and audio captions to uniquely identify the correct time segment(s).\\
    - Require BOTH visual and audio evidence from that event; single modality is INSUFFICIENT.\\
    - Offer plausible options **strictly based on the specific event's details**.\\
    - Be "single\_choice\_question" or "multiple\_choice\_question".\\

    \textbf{2) For answer options:}\\
    - Provide exactly 4 options: A, B, C, D. Each option's value **must be a timestamp string** in "[HH:MM:SS - HH:MM:SS]" format.\\
    - Distractor time segments must be plausible but incorrect for the queried event, ideally from other events or incorrect parts of the correct event.\\
    - For "single\_choice\_question": exactly one correct time segment option.\\
    - For "multiple\_choice\_question": at least two correct time segment options.\\

    \textbf{3) For explanations:}\\
    - Justify the correct answer(s) by synthesizing information **as if you were observing the video directly** and field the `"gold\_reasoning"`.\\
    - Reasoning must detail inference steps, **explicitly referencing the required event\_ids and both visual + audio evidence used to pinpoint the exact time segment.**
\end{tcolorbox}
\caption{Prompt used in QA construction in the \textbf{Multimodal Temporal Localization} task.}
\label{fig:qa-prompt-2}
\vspace{-0.1in}
\end{figure*}

\begin{figure*}[h!]
\begin{tcolorbox}[
    colback=prompt_blue!5!white,
    colframe=frame2,
    title=AUDIO VISUAL ALIGNMENT USER PROMPT,
    fonttitle=\bfseries,
    colbacktitle=prompt_blue!45!white,
    coltitle=frame2!80!white,
    boxrule=1.5pt,
    arc=5pt,
    boxsep=5pt,
    left=12pt,
    right=12pt,
    top=12pt,
    bottom=12pt
]

video\_id: \{video\_id\}\\
summary: \{summary\}\\
events: \{events\_str\}\\

\textbf{Task requirements:}\\
\textbf{1) Generate N=2 **Audio-Visual Alignment** questions.} Each question MUST:\\
- Focus on **identifying the corresponding visual characteristic/expression given an audio event**, \textbf{OR} **identifying the corresponding audio event given a visual characteristic** within a specific event.\\
- Use **exactly 1 event\_id**. The question should target an event's [start\_time] and [end\_time] where the specified audio and visual elements occur concurrently.\\
- Require BOTH visual and audio evidence to correctly identify the aligning characteristic; single modality is INSUFFICIENT.\\
- Offer plausible options **strictly based on the specific event's details**.\\
- Be "single\_choice\_question" or "multiple\_choice\_question".\\

\textbf{2) For answer options:}\\
- Provide exactly 4 options: A, B, C, D. Each option's value **must be a descriptive string** that aligns with the modality being queried (i.e., visual characteristics for visual questions, or audio events for audio questions).\\
- Distractor options must be plausible within the event but not aligned with the queried information, or entirely incorrect.\\
- For "single\_choice\_question": exactly one correct descriptive option.\\
- For "multiple\_choice\_question": at least two correct descriptive options.\\

\textbf{3) For explanations:}\\
- Justify the correct answer(s) by synthesizing information **as if you were observing the video directly** and field the `"gold\_reasoning"`.\\
- Reasoning must detail inference steps, **explicitly referencing the single event\_id and both visual + audio evidence used to align the audio event with its visual manifestation.**

\end{tcolorbox}
\caption{Prompt used in QA construction in the \textbf{Audio-Visual Alignment} task.}
\label{fig:qa-prompt-3}
\vspace{-0.1in}
\end{figure*}

\begin{figure*}[h!]
\begin{tcolorbox}[
    colback=prompt_blue!5!white,
    colframe=frame2,
    title=TIMELINE RECONSTRUCTION USER PROMPT,
    fonttitle=\bfseries,
    colbacktitle=prompt_blue!45!white,
    coltitle=frame2!80!white,
    boxrule=1.5pt,
    arc=5pt,
    boxsep=5pt,
    left=12pt,
    right=12pt,
    top=12pt,
    bottom=12pt
]

video\_id: \{video\_id\}\\
summary: \{summary\}\\
events: \{events\_str\}\\

\textbf{Task requirements:}\\
    \textbf{1) Generate N=2 **Timeline Reconstruction** question.} The question MUST:\\
    - Present a list of 4-10 distinct sub-events in a shuffled, non-chronological order. Each sub-event should be explicitly numbered (e.g., "(1) [Description of sub-event A]", "(2) [Description of sub-event B]"). \\
    - Each sub-event description should be **concise and focuses on a single, atomic action or observation**.\\
    - Sub-events should be drawn from **at least 3 different** event\_ids.\\
    - Require the reconstruction of the correct chronological order of these sub-events.\\
    - Require BOTH visual(e.g., character movements, object appearance/disappearance) and audio(e.g., specific sound effects, spoken time indicators) evidence to determine the correct sequence; single modality is INSUFFICIENT.\\
    - Be "single\_choice\_question".\\

    \textbf{2) For answer options:}\\
    - Provide exactly 4 options: A, B, C, D. Each option's value **must be a sequence of the sub-event numbers**, joined by " -> ".\\
    - Provide exactly 1 correct option which represents the correct chronological sequence of the numbered sub-events.\\
    - Provide exactly 3 distractor options, which must be plausible but incorrect sequences.\\
    - Ensure all sub-event numbers included in the question are used exactly once in each answer option's sequence.\\

    \textbf{3) For explanations:}\\
    - Justify the correct answer(s) by synthesizing information **as if you were observing the video directly** and field the \"gold\_reasoning\".\\
    - Reasoning must detail inference steps, **explicitly referencing the required event\_ids and both visual + audio evidence used to reconstruct the sub-events.**
\end{tcolorbox}
\caption{Prompt used in QA construction in the \textbf{Timeline Reconstruction} task.}
\label{fig:qa-prompt-4}
\vspace{-0.1in}
\end{figure*}

\begin{figure*}[h!]
\begin{tcolorbox}[
    colback=prompt_blue!5!white,
    colframe=frame2,
    title=TOPIC STANCE EVOLUTION SUMMARIZATION USER PROMPT,
    fonttitle=\bfseries,
    colbacktitle=prompt_blue!45!white,
    coltitle=frame2!80!white,
    boxrule=1.5pt,
    arc=5pt,
    boxsep=5pt,
    left=12pt,
    right=12pt,
    top=12pt,
    bottom=12pt
]
video\_id: \{video\_id\}\\
summary: \{summary\}\\
events: \{events\_str\}\\

\textbf{Task requirements:}\\
    \textbf{1) Generate N=2 **Topic/Stance Evolution Summarization** question.} The question MUST:\\
    - Focus on summarizing the **evolution or development of a key topic or a character's stance/viewpoint** across multiple relevant events.\\
    - Involve **at least 3 different event\_ids**.\\
    - Require BOTH visual(e.g., speaker's gestures, on-screen text, changes in setting) and audio(e.g., spoken content, tone shifts, emphasis) evidence to formulate a comprehensive summary; single modality is INSUFFICIENT.\\
    - Offer plausible options **strictly based on the video's main idea**.\\
    - Be "single\_choice\_question" or "multiple\_choice\_question".\\

    \textbf{2) For answer options:}\\
    - Provide exactly 4 options: A, B, C, D. Each option's value **must be a concise, multi-sentence paragraph (2-4 sentences)** describing a potential progression or evolution of the topic/stance across the selected events.\\
    - Distractor options must be plausible descriptions of an evolution, but either not aligned with the actual progression or entirely incorrect.\\
    - For "single\_choice\_question": exactly one correct option.\\
    - For "multiple\_choice\_question": at least two correct options.\\

    \textbf{3) For explanations:}\\
    - Justify the correct answer(s) by synthesizing information **as if you were observing the video directly** and field the \"gold\_reasoning\".\\
    - Reasoning must detail inference steps, **explicitly referencing the required event\_ids and both visual + audio evidence to support the stated progression or evolution.**\\
\end{tcolorbox}
\caption{Prompt used in QA construction in the \textbf{Topic/Stance Evolution Summarization} task.}
\label{fig:qa-prompt-5}
\vspace{-0.1in}
\end{figure*}

\begin{figure*}[h!]
\begin{tcolorbox}[
    colback=prompt_blue!5!white,
    colframe=frame2,
    title=CROSS EVENT CAUSALITY USER PROMPT,
    fonttitle=\bfseries,
    colbacktitle=prompt_blue!45!white,
    coltitle=frame2!80!white,
    boxrule=1.5pt,
    arc=5pt,
    boxsep=5pt,
    left=12pt,
    right=12pt,
    top=12pt,
    bottom=12pt
]
video\_id: \{video\_id\}\\
summary: \{summary\}\\
events: \{events\_str\}\\

\textbf{Task requirements:}\\
    1) Generate N=2 **Cross-event Causality Reasoning** question. The question MUST:\\
    - Choose a specific **"result sub-event"**, which is a localized action or state change within a larger event\_id.\\
    - Ask to identify the preceding event\_id(s) and/or specific sub-event(s) within those event\_id(s) that most plausibly served as the direct cause or primary contributing factor to the target result sub-event.\\
    - The causal relationship must span **at least 3 different event\_ids**.\\
    - Require BOTH visual and audio evidence to robustly establish the causal link; single modality is INSUFFICIENT.\\
    - Offer plausible options **strictly based on the video's main idea**.\\
    - Be "single\_choice\_question" or "multiple\_choice\_question".\\

\textbf{2) For the answer:}\\
    - Provide exactly 4 options: A, B, C, D. Each option should be an event\_ids or a descriptive string of specific sub-events within an event\_id.\\
    - Distractor options must be plausible as preceding events/sub-events, but either not align with the actual causal chain, or are entirely incorrect.\\
    - For "single\_choice\_question": exactly one correct option.\\
    - For "multiple\_choice\_question": at least two correct options.\\

\textbf{3) For explanations:}\\
    - Justify the correct answer(s) by synthesizing information **as if you were observing the video directly** and field the \"gold\_reasoning\".\\
    - Reasoning must detail inference steps, **explicitly referencing the required event\_ids and both visual + audio evidence to support the stated progression or evolution.**\\
    - Clearly explain **how** the referenced cause event(s)/sub-event(s) led to the state change or outcome observed in the target result sub-event.\\
\end{tcolorbox}
\caption{Prompt used in QA construction in the \textbf{Cross-event Causality} task.}
\label{fig:qa-prompt-6}
\vspace{-0.1in}
\end{figure*}

\subsection{Prompts for Scoring QAs}
\label{sec:sco_prompt}
The prompts take the complete visual captions, audio captions, and QA pairs as input, and return a JSON-formatted output containing dimension-wise scores for each QA pair. 
As illustrated in Figure~\ref{fig:sco-prompt}, the scoring prompt is designed to guide GPT-4o in evaluating the initial QA pairs along three dimensions: \textit{sufficiency}, \textit{consistency}, and \textit{relevance}. 

To ensure reliable and fine-grained judgments, the prompt provides explicit scoring criteria, where each dimension is assigned a score between 0 and~1.
Unlike binary filtering, this soft scoring process allows finer discrimination of borderline cases. 
Only QA pairs that achieve a score of 1 across all three dimensions are retained in the final dataset.

\begin{figure*}[h!]
\begin{tcolorbox}[
    colback=prompt_purple!3!white,
    colframe=frame4,
    title=SCORING PROMPT,
    fonttitle=\bfseries,
    colbacktitle=prompt_purple!40!white,
    coltitle=frame4!90!white,
    boxrule=1.5pt,
    arc=5pt,
    boxsep=5pt,
    left=12pt,
    right=12pt,
    top=12pt,
    bottom=12pt
]
You are an evaluator for long audiovisual QA. \\

You will receive three inputs:\\
- Audio caption: a textual description of audio events\\
- Video caption: a textual description of visual events\\
- QA pair: a question and its proposed answer\\

\textbf{Your task:}\\
Evaluate the QA from three perspectives:\\
1. Sufficiency — Do the captions provide enough evidence to support the answer?\\
2. Consistency — Is the answer consistent with the described events (no contradictions)?\\
3. Relevance — Are the captions relevant to the question being asked?\\

\textbf{Scoring:}\\
- Each dimension should be assigned a score between 0 and 1.\\
  * 0 = completely unsupported / inconsistent / irrelevant\\
  * 0.5 = partially supported / somewhat consistent / weakly relevant\\
  * 1 = fully supported / consistent / highly relevant\\

Audio caption: \{audio\_caption\}\\
Video caption: \{video\_caption\}\\
QA pair: \{qa\}
\end{tcolorbox}
\caption{Prompt used for \textbf{scoring QAs}.}
\label{fig:sco-prompt}
\vspace{-0.1in}
\end{figure*}

\end{document}